\documentclass[runningheads]{llncs}
\usepackage{graphicx}

\usepackage{amsthm}
\usepackage{tikz}
\usepackage{comment}
\usepackage{amsmath,amssymb} 
\usepackage{color}
\usepackage{amsmath}
\usepackage{amssymb}
\usepackage[accsupp]{axessibility}  
\usepackage{multirow}
\usepackage{amsfonts}
\usepackage{graphicx}
\usepackage{amsmath}
\usepackage{amsthm}
\usepackage{booktabs}
\usepackage{bm}
\usepackage{algorithm}
\usepackage{algorithmic}

\newcommand{\ie}{\textit{i}.\textit{e}.}
\newcommand{\eg}{\textit{e}.\textit{g}.}
\usepackage[pagebackref,breaklinks,colorlinks]{hyperref}
\usepackage[misc]{ifsym}
\begin{document}
\pagestyle{headings}
\mainmatter
\def\ECCVSubNumber{88}  

\title{InfLoR-SNN: Reducing Information Loss for Spiking Neural Networks} 

\titlerunning{InfLoR-SNN}
%
\author{Yufei Guo\inst{1}\thanks{Equal contribution.} \and Yuanpei Chen\inst{1}$^{\star}$ \and
Liwen Zhang\inst{1} \and YingLei Wang\inst{1} \and Xiaode Liu\inst{1} \and Xinyi Tong\inst{1} \and Yuanyuan Ou\inst{2} \and Xuhui Huang\inst{1} \and Zhe Ma\inst{1}\textsuperscript{\Letter}}
\authorrunning{Guo, Y. et al.}
%
\institute{Intelligent Science \& Technology Academy of CASIC, Beijing 100854, China \\
\and Chongqing University, Chongqing, 400044, China\\
\email{yfguo@pku.edu.cn, rop477@163.com, mazhe\_thu@163.com}} 
\maketitle

\begin{abstract}
    
    The Spiking Neural Network (SNN) has attracted more and more attention recently. It adopts binary spike signals to transmit information. Benefitting from the information passing paradigm of SNNs, the multiplications of activations and weights can be replaced by additions, which are more energy-efficient. However, its ``Hard Reset" mechanism for the firing activity would ignore the difference among membrane potentials when the membrane potential is above the firing threshold, causing information loss. Meanwhile, quantifying the membrane potential to 0/1 spikes at the firing instants will inevitably introduce the quantization error thus bringing about information loss too. To address these problems, we propose to use the ``Soft Reset" mechanism for the supervised training-based SNNs, which will drive the membrane potential to a dynamic reset potential according to its magnitude, and Membrane Potential Rectifier (MPR) to reduce the quantization error via redistributing the membrane potential to a range close to the spikes. Results show that the SNNs with the ``Soft Reset" mechanism and MPR outperform their vanilla counterparts on both static and dynamic datasets.
        
\end{abstract}

\section{Introduction}

Deep Neural Networks (DNNs) have greatly improved many applications in computational vision, \eg, object detection and recognition \cite{2016Deep}, object segmentation \cite{2015U}, object tracking \cite{2016Simple}, etc. In pursuit of models with better performance, more and more complex networks are proposed. However, the increasing complexity poses a new challenge to model deployment on power-constrained devices, thus becoming an impediment to the applications of these advanced complex models. There have been several approaches to address this problem, such as quantization \cite{gong2019differentiable,li2019additive,li2021brecq}, pruning \cite{2017Channel}, knowledge distillation \cite{2018Model}, spiking neural networks (SNNs) \cite{2020Incorporating,2018Direct,li2021free,li2021differentiable,Guo_2022_CVPR,guo2023rmploss,guo2023membrane,guo2023direct,guo2023joint}, and so on. Among these approaches, the biology-inspired method, SNNs provide a unique way to reduce energy consumption by mimicking the spiking nature of brain neurons. A spiking neuron integrates the inputs over time and fires a spike output whenever the membrane potential exceeds the firing threshold. And using 0/1 spike to transmit information makes SNNs enjoy the advantage of multiplication-free inference by converting multiplication to additions. Furthermore, SNNs are energy-efficient on neuromorphic hardwares, such as SpiNNaker \cite{2008SpiNNaker}, TrueNorth \cite{2015TrueNorth}, Darwin \cite{2015Darwin}, Tianjic \cite{2019Towards}, and Loihi \cite{2018Loihi}.

\begin{figure}[t]
	\centering
	\includegraphics[width=0.99\textwidth]{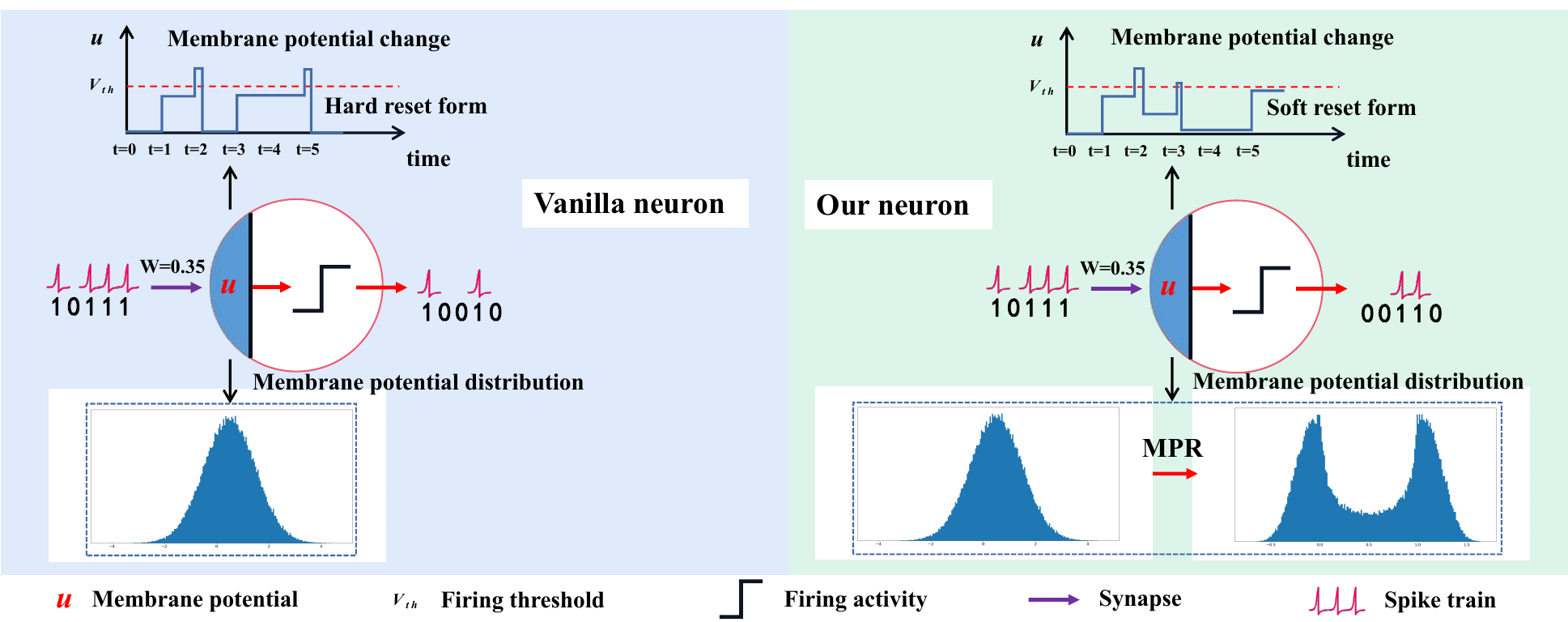} 
	\caption{The difference of our ``Soft Reset"-based neuron and vanilla ``Hard Reset"-based neuron. The membrane potential will be redistributed to reduce the quantization error in our neuron with MPR while not in the vanilla neuron.}
	\label{workflow}
\end{figure}

Despite the attractive benefits, there is still a huge performance gap between existing SNN models and their DNN counterparts. We argue that the reason for the low accuracy is there exists information loss in SNNs. First, the information processing of neurons in supervised training-based SNNs are generally following the rules of the Integrate-and-Fire (IF) model or Leaky IF (LIF) model, where once a membrane potential exceeds the firing threshold, a ``Hard Reset” operation will force the ``residual” potential to be set to $0$, \ie, once fired, all the information will be taken away. Obviously, this mechanism of ``residual” membrane potential-ignored reset mode would fail to preserve the diversity of various membrane potentials. Hence the information encoding capacity of the network is compromised, such that the risk of information loss increases accordingly. Second, although the 0/1 spike information processing paradigm enables SNNs to enjoy the advantage of high efficiency, quantifying the real-valued membrane potential to 0/1 spikes will inevitably introduce the quantization error, which also brings about information loss.

To address the information loss problem, we propose a ``Soft Reset”-based IF (SRIF) neuron model that retains the ``residual” membrane potential from subtracting its spike value at the firing instants. Hence the diversity of the membrane potentials that exceed the firing threshold will be preserved. Though ``Soft Reset” is commonly used in converting methods from ANN to SNN (ANN2SNN)  \cite{2020RMP,2020Deep,li2021free,2019Spiking} methods, 
rarely applied in supervised SNNs~\cite{Ledinauskas2020}, and has not been discussed in SNN enhancement from the perspective of information loss reducing. In addition, for alleviating quantization error, the Membrane Potential Rectifier (MPR) is proposed, which is performed before the firing activity to adjust the membrane potentials towards the spike values (\ie, 0/1). With MPR, the membrane potential will be decoupled as an original one and a modulated one. The original one can keep the mechanism of a neuron and the modulated one enjoys less quantization error than the original one without suffering from any negative effects. The difference between our neuron and the vanilla neuron is illustrated in Fig. \ref{workflow}. Our main contributions are as follows:

\begin{itemize}
	
\item We propose using the SRIF model for supervised training-based SNNs. By retaining the ``residual” membrane potential, SRIF enables the networks to distinguish the differences among those membrane potentials that exceed the firing threshold via subtracting their spike values thus enhancing the information encoding capacity of supervised training-based SNNs.
	
\item We present MPR to mitigate the quantization error. By utilizing a non-linear function to modulate the membrane potential close to 0/1 before firing activity triggers, the gap between the potential and its corresponding 0/1 spike value is minified while maintaining the sparse spike activation mechanism of SNNs. To our best knowledge, few works have noticed the quantization error in SNNs, and a simple but effective method for addressing this problem is presented.

\item Extensive experiments on both static and dynamic datasets were conducted to verify our method. Results show that the SNN trained with the proposed method is highly effective and efficient compared with other state-of-the-art SNN models, \eg, 96.49\% top-1 accuracy and 79.41\% top-1 accuracy are achieved on the CIFAR-10 and CIFAR-100. These results of our models even outperform their DNN counterparts surprisingly, and it is very rare that SNNs may have a chance to surpass their DNN counterparts.

\end{itemize}

\section{Related Work}

\subsection{Learning Methods of Spiking Neural Networks}

The training methods of SNNs can be divided into two categories. The first one is ANN2SNN \cite{2020RMP,2020Deep,li2021free,2019Spiking}. ANN2SNN yields the same input-output mapping for the ANN-SNN pair via approximating the continuous activation values of an ANN using ReLU by averaging the firing rate of an SNN under the rate-coding scheme. Since the ANN has achieved great success in many fields, ANN2SNN can maintain the smallest gap with ANNs in terms of performance and can be generalized to large-scale structures. However, being restricted to rate-coding, ANN2SNN usually requires dozens or even hundreds of timesteps to obtain well-performed networks. Lots of efforts have been done to reduce the long inference time, such as weight normalization \cite{2015Fastclassifying}, threshold rescaling \cite{2019Going}, soft reset \cite{2020RMP}, threshold shift \cite{li2021free}, and the quantization clip-floor-shift activation function \cite{bu2022optimal}, it is still hard to obtain high-performance SNNs with ultra-low latency.

The second one is supervised learning-based SNNs. SNNs quantize the real-valued membrane potentials into 0/1 spikes via the firing activity. Since the gradient of the firing activity function is zero almost everywhere, the gradient descent-based optimizer can not be directly used for the training of SNNs. To alleviate the optimization difficulty, the approximate gradient-based strategy is commonly used, and some related approaches had been proposed to achieve trainable SNNs with high performance. For example, by regarding the SNN as a special RNN, a training method of back-propagation through time with different kinds of surrogate gradient was proposed \cite{2019Surrogate}. The spatio-temporal back-propagation (STBP) \cite{2018Spatio} method enables SNNs to be trained on the ANN programming platform, which also significantly promotes the direct training research of SNNs. Differentiable spike which can match the finite difference gradient of SNNs well was proposed in \cite{li2021differentiable}. The temporal efficient training (TET) \cite{deng2022temporal} method with a novel loss and a gradient descent regime that succeeds in obtaining more generalized SNNs, has also attracted much attention. In RecDis-SNN \cite{Guo_2022_CVPR}, a new perspective to understand the difficulty of training SNNs by analyzing undesired membrane potential shifts is presented and the MPD-Loss to penalize the undesired shifts is proposed. Numerous works verify that supervised learning can greatly reduce the number of timesteps and handle dynamic datasets. It has increasingly aroused researchers’ interest in recent years. In this work, we focus on improving the performance of the supervised learning-based SNNs by repressing information loss, which is rarely mentioned in other works.

\subsection{Threshold-dependent Batch Normalization}

Batch Normalization (BN) is one of the most widely used normalization technologies, which is initially designed for very deep Convolutional Neural Networks (CNNs). As it only focuses on normalizing the spatial feature maps, directly applying BN to SNNs would damage the temporal characteristic of SNNs, which stand with spatio-temporal feature maps, leading to low accuracy. To address this issue, some specially-designed normalization methods for SNNs were proposed recently. Typically, to simultaneously balance neural selectivity and normalize the neuron activity, NeuNorm \cite{2018Spatio} was proposed. Then, a more effective normalization technique that can take good care of the firing threshold, named threshold-dependent Batch Normalization (tdBN) was further proposed in \cite{2020Going}. It can normalize the feature maps of SNNs in both spatial and temporal domains \cite{2020Going}. Specifically, let $\textbf{X}_t \in \mathbb{R}^{B\times C\times H\times W}$ represent the input maps at each timestep, where $t=1,\dots,T$ ($B$: batch size; $C$: channel; $(H, W)$: spatial domain). Then for each channel $c$, the spatio-temporal sequence $\textbf{X}^{(c)} = \{\textbf{X}_1^{(c)}, \cdots ,\textbf{X}_T^{(c)} \}$ is normalized by tdBN as follows,
\begin{equation}\label{tdbn}
	\tilde{\textbf{X}}^{(c)} = \lambda \cdot \frac{\alpha V_{th}(\textbf{X}^{(c)}-\bar{x}^{(c)})}{\sqrt{{\rm mean}((\textbf{X}^{(c)}-\bar{x}^{(c)})^2)+\epsilon}} + \beta,
\end{equation}
where $V_{th}$ is the firing threshold, $\alpha$ is a network-structure-dependent hyper-parameter, $\epsilon$ is a tiny constant, $\lambda$ and $\beta$ are two learnable parameters, $\bar{x}^{(c)}={\rm mean}(\textbf{X}^{(c)})$ is the mean value of $\textbf{X}^{(c)}$, $\tilde{\textbf{X}}^{(c)}$ is the normalized maps. In this paper, tdBN is also adopted considering its spatio-temporal normalization mechanism.

\section{Preliminary and Methodology}

To avoid the information loss in supervised training-based SNNs, we propose the ``Soft Reset'' IF (SRIF) model and Membrance Potential Rectificater (MPR).

\subsection{``Soft Reset" IF Model}
An SNN adopts a biology-inspired spiking neuron that accumulates inputs along the time dimension as its membrane potential and fires a spike when the potential exceeds the firing threshold. This mechanism makes it much different from its DNN counterpart. For better introducing the proposed SRIF neuron, a unified form defined by a recent work \cite{2020Incorporating}, is given to describe the dynamics of all kinds of spiking neurons as follows,
\begin{equation}\label{ht11}
	H[t] = f(U[t-1],X[t]),
\end{equation}
\begin{equation}\label{ot}
	O[t] = \Theta(H[t]-V_{th}),
\end{equation}
\begin{equation}\label{vt}
	U[t] = H[t](1-O[t])+V_{reset}O[t],
\end{equation}
where $X[t]$, $H[t]$, $U[t]$, and $O[t]$ are the input, membrane potentials before and after the trigger of a spike, and output spike at the timestep $t$, respectively. $V_{th}$ is the firing threshold, and is usually set to $0.5$. $\Theta(\cdot)$ is the step function defined by $\Theta(x) = 1$ for $x \ge 0$ and $\Theta(x) = 0$ for $x < 0$. $V_{reset}$ denotes the reset potential, which is set as $0$. The function $f(\cdot)$ describes the neuronal dynamics of spiking neuron models, for the commonly used IF neuron and LIF neuron, $f(\cdot)$ can be respectively defined as follows, 
\begin{equation}\label{ht}
	H[t] = U[t-1]+X[t],
\end{equation}
\begin{equation}\label{ht}
	H[t] = \tau U[t-1]+ X[t],
\end{equation}
where $\tau$ denotes the membrane time constant. 

Both LIF and IF neurons have some unique advantages, with decay characteristics introduced by the membrane time constant, LIF neuron behaves more biologically compared with IF neuron, while IF neuron is more efficient due to its addition-only processing manner. In terms of accuracy performance, neither of them show an overwhelming advantage, and more detailed experimental results of these two neurons are provided in Section 4. Considering the subtle gap in performance, we prefer to use LIF model due to its neurodynamic characteristic, from the perspective of brain science research. Conversely, from the perspective of computer science research, we recommend using IF model, since it is more friendly to hardwares.

However, both the IF model and LIF model might undertake a greater or lesser risk of information loss by the ``Hard Reset" mechanism, \ie, when the input membrane potentials exceed the firing threshold, the neurons will  force the membrane potentials to a fixed value. Such mechanism ignores the ``residual" parts of those fired membrane potentials. These ``residual" parts contain the diversity of the input potentials, and we argue that a neuron model which can preserve the diversity or differences of these membrane potentials that cause the firing is more suitable. 

\begin{figure}[t]
	\centering
	\includegraphics[width=0.95\textwidth]{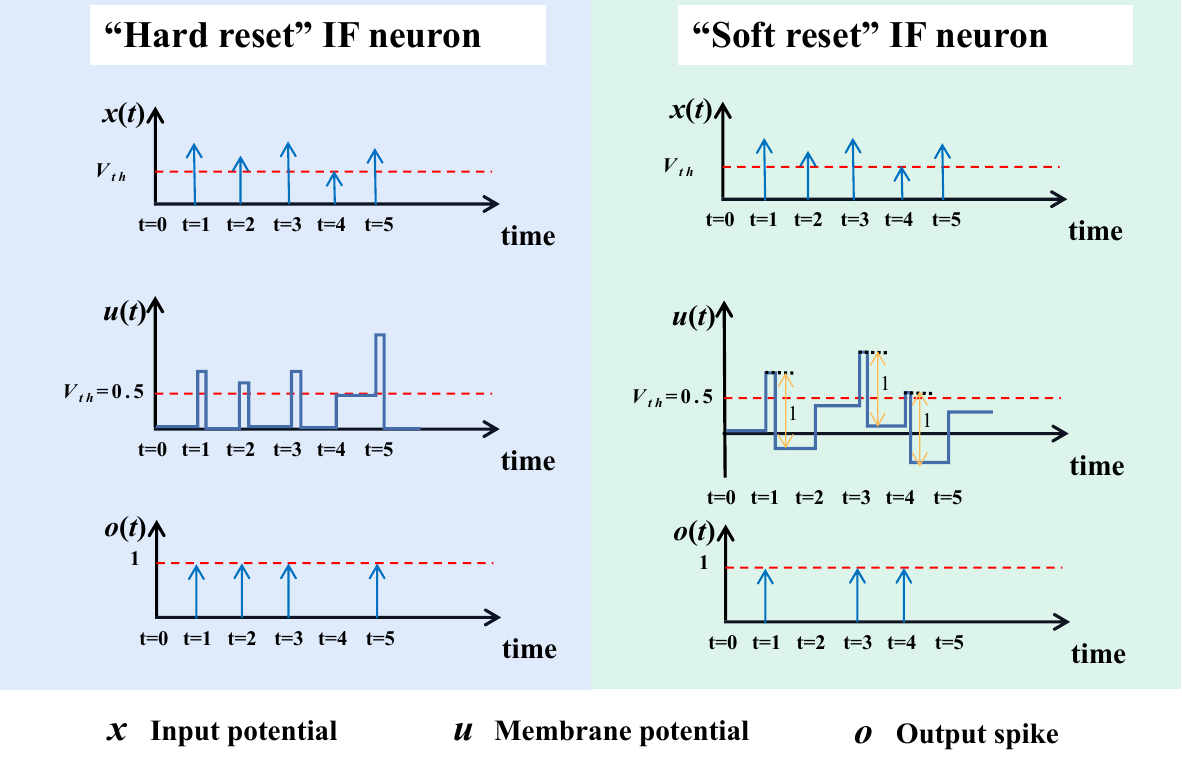} 
	\caption{The difference of ``Hard Reset” IF neuron and ``Soft Reset” IF (SRIF) neuron.}
	\label{softreset}
\end{figure}

To this end, along with the consideration of efficiency, we propose using a ``Soft Reset" mechanism-based IF neuron, SRIF, which can keep the diversity of the membrane potentials by subtracting their firing spike values from themselves at the time where the threshold is exceeded. Though this similar ``Soft Reset” mechanism has been widely used in ANN2SNN \cite{2020RMP,2020Deep,li2021free,2019Spiking}, there are few works to use it in supervised learning-based SNNs~\cite{Ledinauskas2020}. We found 
its value in this field from a new perspective to reduce information loss.
In SRIF neuron, Eq. \eqref{vt} is updated as
\begin{equation}\label{svt}
	U[t] = H[t](1-O[t])+(H[t]-O[t])O[t].
\end{equation}
It can be further simplified as
\begin{equation}\label{svt2}
	U[t] = H[t]-O[t].
\end{equation}
It can be seen that, similar to IF neuron, SRIF is also an addition-only model, thus enjoying computational efficiency when implementing on hardwares. Fig. \ref{softreset} compares the difference between IF neuron and SRIF neuron in an intuitive way. Suppose that both models receive weighted input sequence of 1.5$V_{th}$, 1.2$V_{th}$, 1.5$V_{th}$, 0.9$V_{th}$, and 1.4$V_{th}$ across $5$ consecutive timesteps. Our SRIF neuron will produce three spikes by retaining the residual potentials at the firing instants as depicted in Fig. \ref{softreset}. Whereas, the IF neuron will produce four spikes.

\subsection{Membrane Potential Rectificater}

To further mitigate the information loss, we present a non-linear function, called MPR by reducing the quantization error. MPR aims to redistribute the membrane potential before it is operated by the step function. It only modulates the membrane potential that is presented to the step function but does not modify the value of membrane potential, which receives and accumulates spikes from other neurons. Specifically, we further distinguish the membrane potentials as the original one, ${H}$ as in Eq. \eqref{ht11} and the modulated one, $\hat{H}$, which is the membrane potential that will be presented to the step function. In all previous works, ${H}$ and $\hat{H}$ are treated as the same. While in this paper, we would like to provide a new perspective that using a decoupling function to separate ${H}$ and $\hat{H}$ can be helpful. Specifically, $H$ manages the original tasks as in other work, $\hat{H}$ derives from $H$ with a non-linear function, $\varphi(\cdot)$, and it will be fed into the step function with a modulated form that can shrink the quantization error. With this decoupling mechanism, a neuron model can not only keep the membrane potential updating rule but also enjoy less quantization error.

Before giving the full details of the MPR, we try to formulate the quantization error first. It is clear that the quantization errors corresponding to different membrane potentials should be different. Hence, a value closer to its quantization spike, $o$, enjoys less quantization error. In specific, the firing threshold divides the membrane potentials into two parts, the part with smaller values is assigned to ``$0$" spike, and the other with larger values is assigned to ``$1$" spike. Then the quantization error depends on the margin between the membrane potential and its corresponding spike. Therefore, the quantization error can be defined as the square of the difference between the membrane potential and its corresponding quantization spike value as follows:

\begin{equation}\label{err1}
	\mathcal{L}_q =  (u-o)^2,
\end{equation}
where $u$ is the membrane potential and $o \in \{0,1\}$. when $u$ is below the firing threshold, $o$ is $0$, otherwise, $1$.




\begin{figure}[t]
	\centering
	\includegraphics[width=0.50\textwidth]{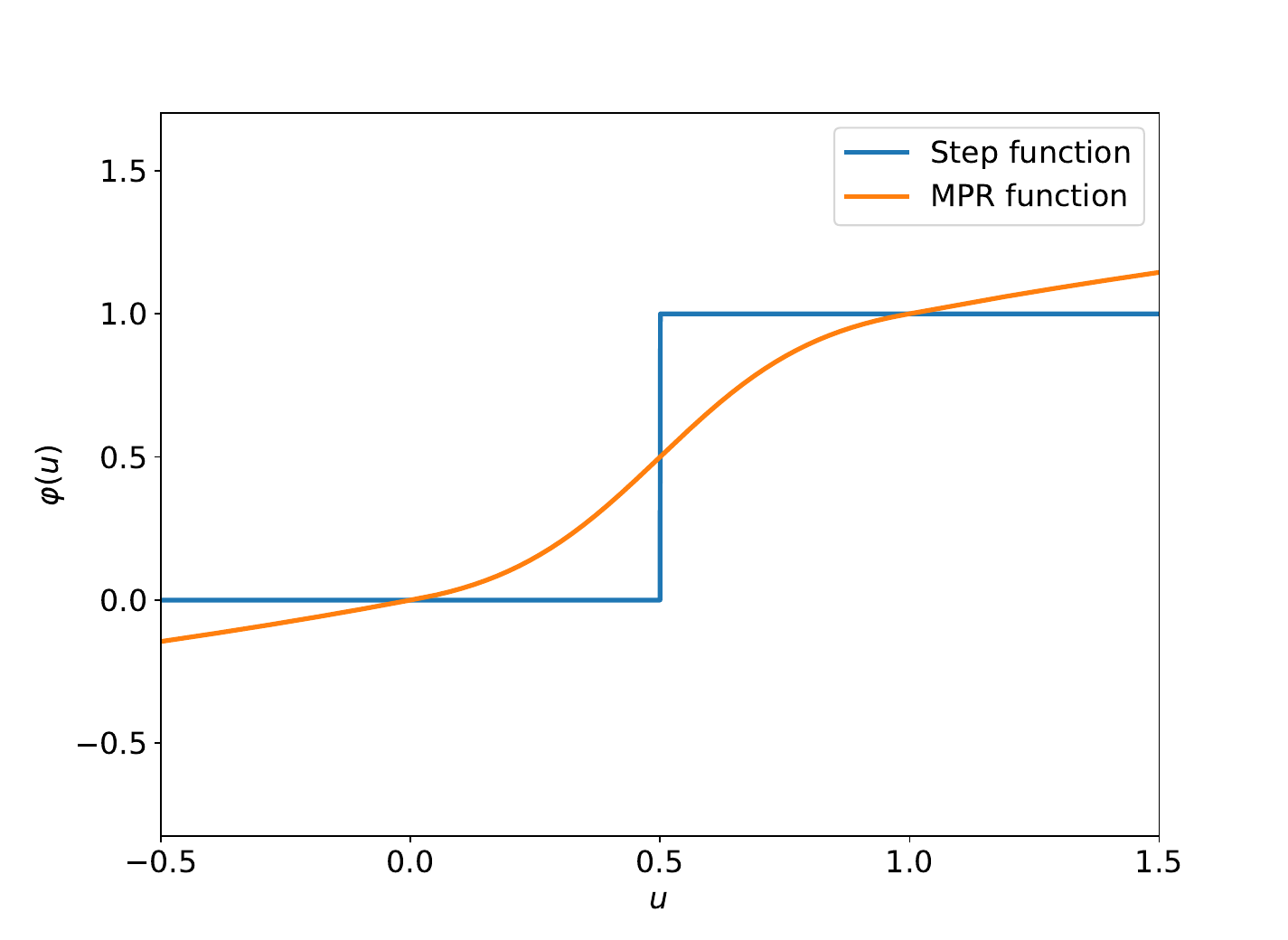} 
	\caption{The MPR function.}
	\label{mprfuntion}
\end{figure}

Hence, the design of MPR should obey the following two principles:

\begin{itemize}
	
\item \textbf{Spike-approaching}: the modulated membrane potential, $\hat{H}$ should be closer to the $0/1$ spikes than the original membrane potential, ${H}$. This principle ensures quantization error reduction.
	
\item \textbf{Firing-invariance}: for the ${H}$ less than $V_{th}$, the MPR should not produce the $\hat{H}$ greater than $V_{th}$ and vice versa. This principle ensures the neuron output be consistent with or without using MPR.

\end{itemize}

Based on the above two principles, we define the MPR as the following symmetrical function:
\begin{equation}\label{mpr}
\varphi (u) =
\left\{
     \begin{array}{ll}
     -(1-u)^{1/3}+1, & u \textless 0,  \\
     \frac{1}{2tanh(3/2)} tanh(3(u-1/2))+1/2, & 0\leq u\leq 1,\\
     (u)^{1/3}, & u \textgreater 1.
     \end{array}
\right.
\end{equation}
Fig. \ref{mprfuntion} shows the response curve of the designed MPR function following the principles of spike-approaching and firing-invariance.

\begin{figure}[t]
	\centering
	\includegraphics[width=0.9\textwidth]{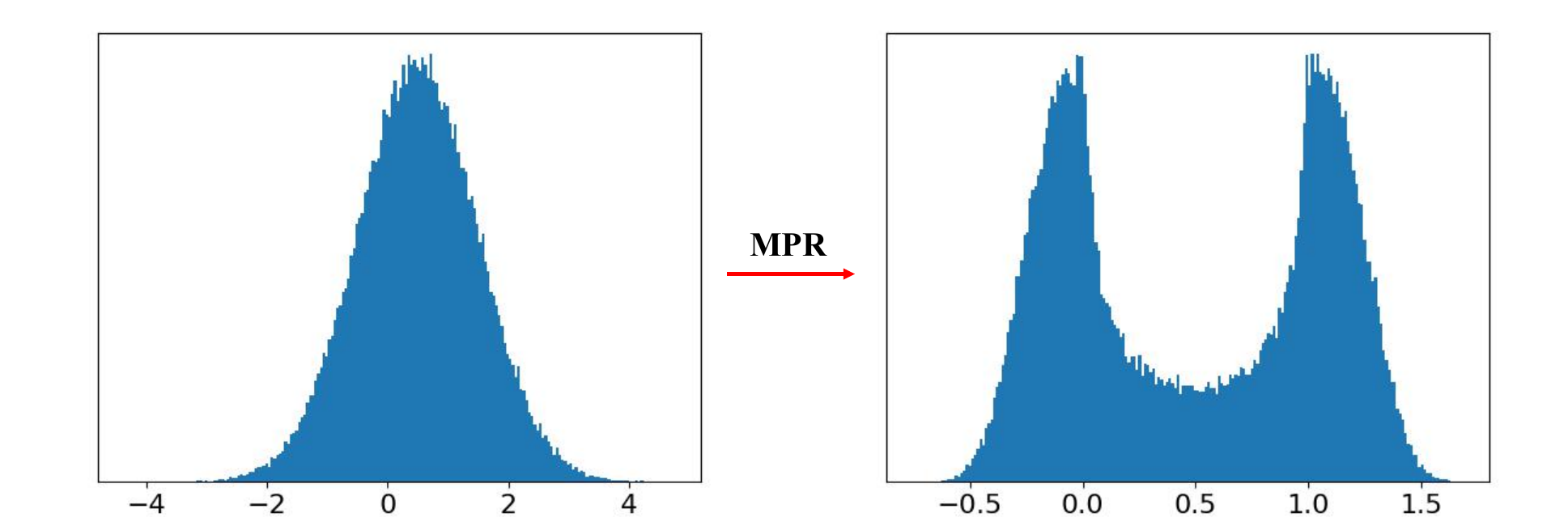} 
	\caption{The effect of the MPR. The original membrane potential distribution (left). The redistributed membrane potential distribution by MPR (right).}
	\label{mpreffect}
\end{figure}

According to \cite{2020Going}, the membrane potential follows a Gaussian distribution, $\mathcal{N}(\mu ; \sigma)$. Hence, to visualize the effect of the MPR, we sample 1000,00 values from a Gaussian distribution with $\mathcal{N}(1/2 ; 1)$, and present them to the MPR. Then the distribution of these 1000,00 MPR outputs is drawn in Fig. \ref{mpreffect}. It can be seen that the unimodal distribution, $\mathcal{N}(1/2 ; 1)$ is adjusted to a bimodal distribution which is with less quantization error since it can naturally gather the membrane potentials near ``0" and ``1".

Moreover, it is worth noting that, the redistributed membrane potential, $\hat{H}$ by MPR is only used for narrowing the gap between the true membrane potential, ${H}$ and its quantization spike. It will not replace the original ${H}$ in our SRIF neuron model. Then the complete new dynamics of the SRIF model can be described as follows,

\begin{equation}\label{srif1}
	H[t] = U[t-1]+X[t],
\end{equation}
\begin{equation}\label{srif2}
	\hat{H}[t] = \varphi{(H[t])},
\end{equation}
\begin{equation}\label{srif3}
	O[t] = \Theta(\hat{H}[t]-V_{th}),
\end{equation}
\begin{equation}\label{srif4}
	U[t] = H[t]-O[t].
\end{equation}

The detailed Feed-Forward procedure for the SRIF neuron with MPR is given in Algo.1.
\begin{algorithm}[tb]
	\caption{Feed-Forward procedures for the ``soft reset" IF neuron with MPR.}
	\label{alg:iresg}
	\textbf{Input}: the input current, $X$.\\
	\textbf{Output}: the output spike train, $O$.
	
	\textbf{Feed-Forward}:
	\begin{algorithmic}[1] 
		\FOR {for all $t =  1, 2, \dots , T\text{-th}$ timesteps}
		\STATE Update the membrane potential, $H(t)$ by Eq. \eqref{srif1}, which represents the membrane potential accumulating the input current.
		\STATE Redistribute the membrane potential, $H(t)$ by Eq. \eqref{srif2} and denote the redistributed membrane potential as $\hat{H}[t]$.
		\STATE Calculate the output spike, $O(t)$ by Eq. \eqref{srif3} using the new membrane potential, $\hat{H}[t]$.
		\STATE Update the membrane potential, $U(t)$ by Eq. \eqref{srif4}, which represents the membrane potential after the trigger of a spike.
		\ENDFOR
	\end{algorithmic}
\end{algorithm}

\section{Experiment}

The proposed methods were evaluated on various static datasets (CIFAR-10~\cite{CIFAR-10}, CIFAR-100~\cite{CIFAR-10}, ImageNet~\cite{2009ImageNet}) and one neuromorphic dataset (CIFAR10-DVS~\cite{2017CIFAR10}) with widely-used spiking archetectures including ResNet20~\cite{2020DIET,2019Going}, VGG16~\cite{2020DIET}, ResNet18~\cite{2021Deep},  ResNet19~\cite{2020Going}, and ResNet34~\cite{2021Deep}.

\subsection{Datasets and Settings}

\textbf{Datasets.} The CIFAR-10(100) dataset consists of 60,000 images in 10(100) classes with $32\times 32$ pixels. The number of the training images is 50,000, and that of the test images is 10,000. The CIFAR10-DVS dataset is the neuromorphic version of the CIFAR-10 dataset. It is composed of 10,000 images in 10 classes, with 1000 images per class. ImageNet dataset has more than
1,250,000 training images and 50,000 test images.

\textbf{Preprocessing.} Data normalization is applied on all static datasets to ensure that input images have $0$ mean and $1$ variance. Besides, the random horizontal flipping and cropping on these datasets were conducted to avoid overfitting. For CIFAR-10, the  AutoAugment~\cite{cubuk2019autoaugment} and  Cutout~\cite{devries2017improved} were used for data augmentation. For the neuromorphic dataset, since the CIFAR10-DVS dataset does not separate data into training and testing sets, we split the dataset into 9000 training images and 1000 test images similar to~\cite{2018Direct}. For data preprocessing and augmentation, we resized the training image frames to $48\times 48$ as in~\cite{2020Going} and adopted random horizontal flip and random roll within $5$ pixels. And the test images are just resized to $48\times 48$ without any additional processing. 

\textbf{Training setup.} For all the datasets, the firing threshold $V_{th}$ was set as $0.5$ and $V_{reset}$ as $0$. For static image datasets, the images were encoded to binary spike using the first layer of the SNN, as in recent works~~\cite{2020DIET,2020Incorporating,2021Deep}. This is similar to rate-coding. For the neuromorphic image dataset, we used the $0/1$ spike format directly. The neuron models in the output layer accumulated the incoming inputs without generating any spike as the output like in~\cite{2020DIET}. For CIFAR-10(100) and CIFAR10-DVS datasets, the SGD optimizer with the momentum of $0.9$ and learning rate of $0.01$ with cosine decayed~\cite{2016SGDR} to $0$. All models were trained within $400$ epochs with the same batch size of $128$. For the ImageNet dataset, the SGD optimizer with a momentum set as $0.9$ and a learning rate of $0.1$ with cosine decayed~\cite{2016SGDR} to 0. All models are trained within $320$ epochs as in~\cite{2021Deep}. The batch size is set to $64$. 

\begin{table}[t]
	\centering	
	\caption{Ablation study for different neuron models without MPR.}	
	\label{tab:ab}
	 \setlength{\tabcolsep}{5mm}{
	\begin{tabular}{llcc}	
		\toprule
		Dataset & Neuron model & Timestep & Accuracy \\	
		\toprule
		\multirow{12}{*}{CIFAR-10}	
		& ``Hard Reset" LIF & 2 & 90.36\%   \\	
		& ``Hard Reset" IF & 2 & 90.07\%   \\
		& ``Soft Reset" IF (SRIF) & 2 & \textbf{90.38\%}   \\
		\cline{2-4}
		& ``Hard Reset" LIF & 4 & 92.22\%   \\	
		& ``Hard Reset" IF & 4 & 92.04\%   \\
		& ``Soft Reset" IF (SRIF) & 4 & \textbf{92.46\%}   \\
		\cline{2-4}
		& ``Hard Reset" LIF & 6 & 92.66\%   \\	
		& ``Hard Reset" IF & 6 & 92.26\%   \\
		& ``Soft Reset" IF (SRIF) & 6 & \textbf{93.40\%}   \\
		\cline{2-4}
		& ``Hard Reset" LIF & 8 & 92.90\%   \\	
		& ``Hard Reset" IF & 8 & 92.86\%   \\
		& ``Soft Reset" IF (SRIF) & 8 & \textbf{94.09\%}   \\
		\hline
		\multirow{12}{*}{CIFAR-100}	
		& ``Hard Reset" LIF & 2 & 62.67\%   \\	
		& ``Hard Reset" IF & 2 & 63.43\%   \\
		& ``Soft Reset" IF (SRIF) & 2 & \textbf{63.85\%}   \\
		\cline{2-4}
		& ``Hard Reset" LIF & 4 & 66.00\%   \\	
		& ``Hard Reset" IF & 4 & 66.95\%   \\
		& ``Soft Reset" IF (SRIF) & 4 & \textbf{67.90\%}   \\
		\cline{2-4}
		& ``Hard Reset" LIF & 6 & 67.44\%   \\	
		& ``Hard Reset" IF & 6 & 68.31\%   \\
		& ``Soft Reset" IF (SRIF) & 6 & \textbf{69.59\%}   \\
		\cline{2-4}
		& ``Hard Reset" LIF & 8 & 67.85\%   \\	
		& ``Hard Reset" IF & 8 & 69.14\%   \\
		& ``Soft Reset" IF (SRIF) & 8 & \textbf{69.90\%}   \\
		\bottomrule			   		         	            			         	
	\end{tabular}
	}
\end{table}

\subsection{Ablation Study for Different Neuron Models}

We first conducted a set of ablation experiments to verify the effectiveness of the proposed SRIF model on CIFAR-10(100) using ResNet20 as the backbone under various timesteps without MPR. The results are shown in Tab. 1.

It can be seen that whether on CIFAR-10 or CIFAR-100, the SRIF neuron always obtains the best result ranging from 2 timesteps to 8 timesteps. This indicates the superiority of the SRIF neuron. On the other hand, the LIF neuron performs better than the ``Hard Reset" IF neuron on CIFAR-10, while the IF neuron performs better on CIFAR-100, even though the LIF neuron is more like a biological neuron. This comparison also shows that, although SNNs are proposed to imitate the biological neural networks, for the implementation of large-scale networks, they still need to rely on computer hardwares. Hence, the characteristics of computational science should also be considered. In this respect, the SRIF neuron is more suitable for its advantage of low power consumption and capacity of reducing information loss.

\begin{table}[tp]
	\centering	
	\caption{Ablation study for MPR.}	
	\label{tab:ab}
	 \setlength{\tabcolsep}{2.5mm}{
	\begin{tabular}{lllcc}	
		\toprule
		Dataset & Architecture & Method & Timestep & Accuracy \\	
		\toprule
		\multirow{4}{*}{CIFAR-10} & \multirow{2}{*}{ResNet20} & SRIF w/o MPR & 4 & 92.46\%   \\	
		                          &                           & SRIF w/ MPR & 4 & \textbf{92.94\%}   \\
		\cline{2-5}
		                          &\multirow{2}{*}{ResNet19}  & SRIF w/o MPR & 4 & 95.44\%   \\	
		                          &                           & SRIF w/ MPR & 4 & \textbf{96.27\%}   \\
		\hline
		\multirow{4}{*}{CIFAR-100}& \multirow{2}{*}{ResNet20} & SRIF w/o MPR & 4 & 67.90\%   \\	
		                          &                           & SRIF w/ MPR & 4 & \textbf{70.63\%}   \\	
		\cline{2-5}
		                          &\multirow{2}{*}{ResNet19}  & SRIF w/o MPR & 4 & 77.85\%   \\	
		                          &                           & SRIF w/ MPR & 4 & \textbf{78.42\%}   \\		
		\bottomrule			   		         	            			         	
	\end{tabular}
	}
\end{table}

\subsection{Addition of MPR}

\begin{table}[tp]
	\centering	
	\caption{ Quantization error.}	
	\label{tab:ab}
	 \setlength{\tabcolsep}{2.5mm}{
	\begin{tabular}{lllcc}	
		\toprule
		Dataset & Architecture & Method & Timestep & Avg. error \\	
		\toprule
		\multirow{4}{*}{CIFAR-10} &  \multirow{2}{*}{ResNet20} & Before MPR & 4 & 0.28   \\	
		                          &                            & After MPR & 4 & \textbf{0.04}   \\
		\cline{2-5}
		                          &\multirow{2}{*}{ResNet19}  & Before MPR & 4 & 0.20   \\	
		                          &                           & After MPR & 4 & \textbf{0.03}   \\		
		\hline
		\multirow{4}{*}{CIFAR-100}& \multirow{2}{*}{ResNet20} & Before MPR & 4 & 0.38   \\	
		                          &                           & After MPR & 4 & \textbf{0.05}   \\
		\cline{2-5}
		                          &\multirow{2}{*}{ResNet19}  & Before MPR & 4 & 0.32   \\	
		                          &                           & After MPR & 4 & \textbf{0.04}   \\			
		\bottomrule			   		         	            			         	
	\end{tabular}
	}
\end{table}

\begin{figure}[h]
	\centering
	\includegraphics[width=0.62\textwidth]{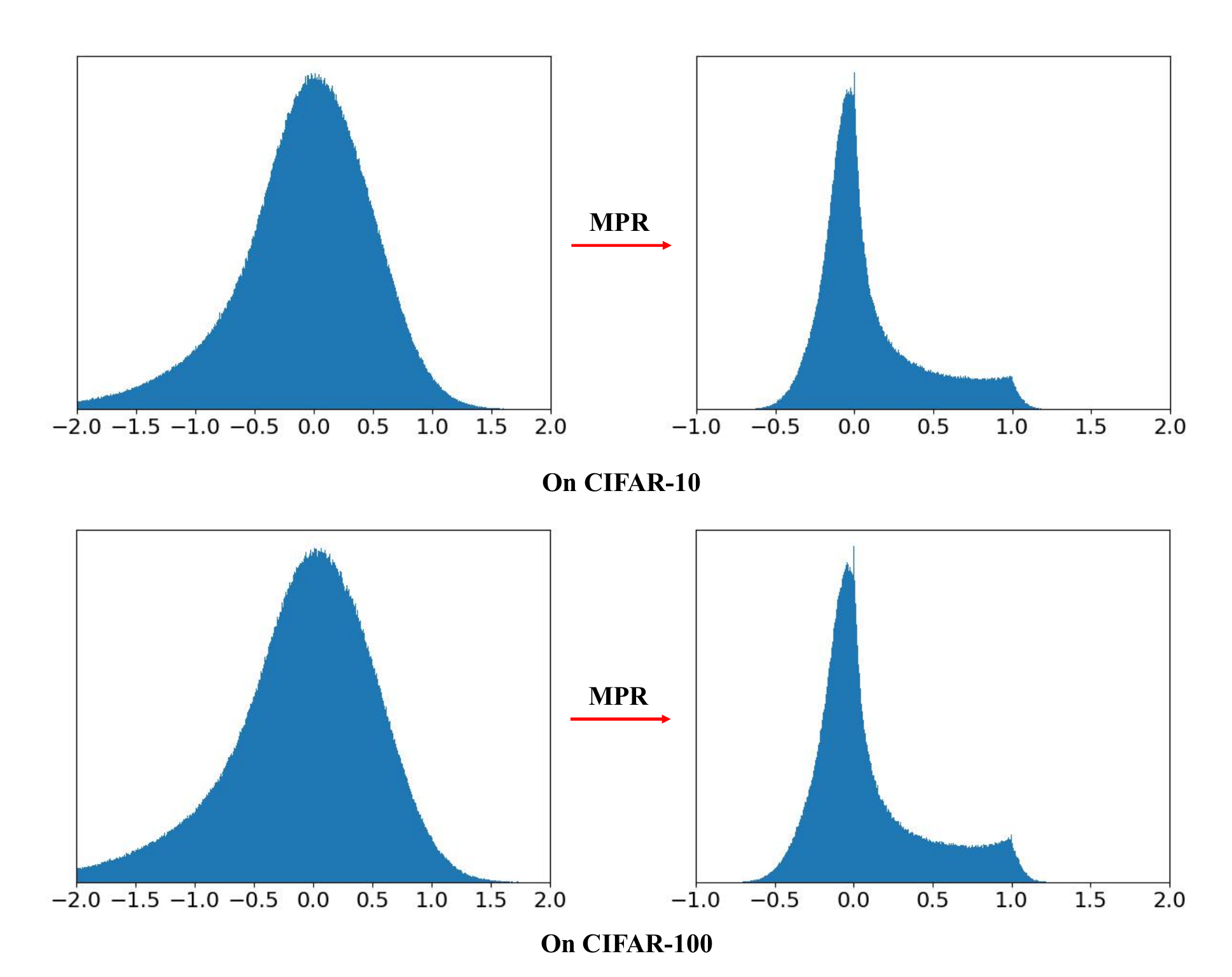} 
	\caption{The effect of MPR. The overall original membrane potential distribution (left) and the redistributed membrane potential distribution by MPR (right) of the first layer of the second block in ResNet20 on CIFAR-10 and CIFAR-100 test sets.}
	\label{distribution}
\end{figure}
Then, a set of ablation experiments for the MPR were conducted on CIFAR-10(100) using ResNet20 and ResNet19 as backbones within $4$ timesteps. Results in Tab. 2 show that the MPR can greatly improve performance. Especially on CIFAR-100, where ResNet20 with MPR increases the accuracy by $2.73\%$. These results verify the effectiveness of MPR in terms of performance improvement.

We also computed the average quantization error of the first layer of the second block in the ResNet20/19 before and after MPR on the test set of CIFAR-10(100), respectively. Results in Tab. 3 show that the quantization error is obviously reduced by the MPR. The overall original membrane potential distribution and modulated membrane potential distribution by MPR of the first layer of the second block in ResNet20 on CIFAR-10 and CIFAR-100 test sets are shown in Fig. \ref{distribution}. It shows that the MPR adjusts the membrane potential distribution near ``$0$" and ``$1$", which is closer to its quantization spike. Put together, these results quantitatively support the effectiveness of MPR in reducing quantization error.

\subsection{Comparisons with Other Methods}

\begin{table*}[tp]\small
	\centering	
	\caption{Comparison with SoTA methods.* denotes self-implementation results.}	
	\label{tab:Comparison}	
	\begin{tabular}{lllccc}	
		\toprule
		Dataset & Method & Type & Architecture & Timestep & Accuracy \\	
		\toprule
		\multirow{21}{*}{CIFAR-10}	
		& SpikeNorm~\cite{2019Going} & ANN2SNN & VGG16 & 2500 & 91.55\%   \\	
		& Hybrid-Train~\cite{2020Enabling} & Hybrid & VGG16 & 200 & 92.02\%   \\
		& Spike-basedBP~\cite{2019Enabling} & SNN training & ResNet11 & 100 & 90.95\%   \\
		& STBP~\cite{2018Direct} & SNN training & CIFARNet & 12 & 90.53\%   \\  	
		& TSSL-BP~\cite{2020Temporal} & SNN training & CIFARNet & 5 & 91.41\%   \\ 
		& PLIF~\cite{2020Incorporating} & SNN training & PLIFNet & 8 & 93.50\%   \\
		\cline{2-6}
		& \multirow{4}{*}{Diet-SNN~\cite{2020DIET}} & \multirow{4}{*}{SNN training} & \multirow{2}{*}{VGG16} & 5 & 92.70\%   \\ 
		& &  &  & 10 & 93.44\%   \\ 
		\cline{4-6}
		&  &  & \multirow{2}{*}{ResNet20} & 5 & 91.78\%   \\ 
		&  &  &  & 10 & 92.54\%   \\   
		\cline{2-6}
		& \multirow{3}{*}{STBP-tdBN~\cite{2020Going}} & \multirow{3}{*}{SNN training} & \multirow{3}{*}{ResNet19} 
		& 2 & 92.34\%   \\
		&  &  &											                                  & 4 & 92.92\%   \\
		&  &  &											                                   & 6 & 93.16\%   \\
		\cline{2-6}
		& ANN*  & ANN & ResNet19 & 1 & 96.29\%   \\
		\cline{2-6}
		& \multirow{7}{*}{\textbf{InfLoR-SNN}} & \multirow{7}{*}{SNN training} & \multirow{3}{*}{ResNet19} 
		& 2 & \textbf{94.44\%}$\pm 0.08$  \\
		&  &  &											                                  & 4 & \textbf{96.27\%}$\pm 0.07$  \\
		&  &  &											                                   & 6 & \textbf{96.49\%}$\pm 0.08$   \\	
		\cline{4-6}	
		&  &  & \multirow{2}{*}{ResNet20} 		                                          & 5 & \textbf{93.01\%}$\pm 0.06$   \\
		&  &  &											                                  & 10 & \textbf{93.65\%}$\pm 0.04$   \\
		\cline{4-6}		
		&  &  &	\multirow{2}{*}{VGG16}                                                   & 5 & \textbf{94.06\%}$\pm 0.08$  \\
		&  &  &											                                 & 10 & \textbf{94.67\%}$\pm 0.07$   \\		
		\hline	
		
		\multirow{14}{*}{CIFAR-100}	
		& BinarySNN~\cite{2020Exploring} & ANN2SNN & VGG15 & 62 & 63.20\%   \\
		& Hybrid-Train~\cite{2020Enabling} & Hybrid & VGG11 & 125 & 67.90\%   \\  	
		& T2FSNN~\cite{2020T2FSNN} & ANN2SNN & VGG16 & 680 & 68.80\%   \\ 
		& Burst-coding~\cite{park2019fast} & ANN2SNN & VGG16 & 3100 & 68.77\%   \\
		& Phase-coding~\cite{Kim2018Deep} & ANN2SNN & VGG16 & 8950 & 68.60\%   \\
		\cline{2-6} 
		& \multirow{2}{*}{Diet-SNN~\cite{2020DIET}} & \multirow{2}{*}{SNN training} & ResNet20 & 5 & 64.07\%   \\ 
		&                                          &                                & VGG16 & 5 & 69.67\%   \\  
		\cline{2-6}
		& ANN*  & ANN & ResNet19 & 1 & 78.61\%   \\
		\cline{2-6}
		& \multirow{6}{*}{\textbf{InfLoR-SNN} } & \multirow{6}{*}{SNN training} & {ResNet20} 
		& 5 & \textbf{71.19\%}$\pm 0.09$   \\
		\cline{4-6}	
		&  &  &	\multirow{2}{*}{VGG16}                                                       & 5 & \textbf{71.56\%}$\pm 0.10$   \\
		&  &  &											                                     & 10 & \textbf{73.17\%}$\pm 0.08$   \\
		\cline{4-6}	
		&  &  &	\multirow{3}{*}{ResNet19}                                                    & 2 & \textbf{75.56\%}$\pm 0.11$  \\
		&  &  &											                                     & 4 & \textbf{78.42\%}$\pm 0.09$  \\
		&  &  &											                                     & 6 & \textbf{79.51\%}$\pm 0.06$   \\	
		\hline		
		\multirow{9}{*}{ImageNet}	
		& Hybrid-Train~\cite{2020Enabling} & Hybrid & ResNet34 & 250 & 61.48\%   \\  
		& SpikeNorm~\cite{2019Going} & ANN2SNN & ResNet34 & 2500 & 69.96\%   \\		
		& STBP-tdBN~\cite{2020Going} &  SNN training & ResNet34 & 6 & 63.72\%   \\ 
		\cline{2-6}
		& \multirow{2}{*}{SEW ResNet~\cite{2021Deep}} & \multirow{2}{*}{SNN training} & {ResNet18} & 4 & {63.18\%}   \\
		&  &  &											                             {ResNet34} & 4 & {67.04\%}   \\ 
		\cline{2-6}
		& \multirow{2}{*}{Spiking ResNet~\cite{2021Deep}} & \multirow{2}{*}{SNN training} & {ResNet18} & 4 & {62.32\%}   \\
		&  &  &											                             {ResNet34} & 4 & {61.86\%}   \\ 
		\cline{2-6}
		& \multirow{2}{*}{\textbf{InfLoR-SNN} } & \multirow{2}{*}{SNN training} & {ResNet18} & 4 & \textbf{64.78\%}$\pm 0.07$   \\
		&  &  &											                             {ResNet34} & 4 & {65.54\%}$\pm 0.08$   \\
		\bottomrule				         	
	\end{tabular}	
\end{table*}

Our method was further compared with other state-of-the-art SNNs on static and neuromorphic datasets. Results are shown in Tab. 4, where for each run, the mean accuracy and standard deviation of 3 trials are listed. For simplification, \textbf{InfLoR} (\ie, short for \textbf{Inf}ormation \textbf{Lo}ss \textbf{R}educing) is used to denote the combination of SRIF and MPR.

\textbf{CIFAR-10(100).}
For CIFAR-10, our method improves network performance across all commonly used backbones in SNNs. ResNet19-based InfLoR-SNN achieved 96.49\% top-1 accuracy with 6 timesteps, which outperforms its STBP-tdBN counterpart with 3.33\% higher accuracy and its ANN counterpart 0.20\% higher accuracy even. The ResNet20-based InfLoR-SNN can reach to 93.65\%, while only 92.54\% in \cite{2020DIET}. And our VGG16-based network also shows higher accuracy than other methods with fewer timesteps. On CIFAR-100, InfLoR-SNN also performs better and achieves a 1.89\% increment on VGG16. Noteworthy, InfLoR-SNN significantly surpasses Diet-SNN \cite{2020DIET} with 7.12\% higher accuracy, which is not easy to achieve in the SNN field. Again, our ResNet19 also outperforms its ANN counterpart. To our best knowledge, it is the first time that the SNN can outperform its ANN counterpart.

\textbf{ImageNet.}
For the ImageNet dataset, ResNet18 and ResNet34 were used as the backbones. Results show that our ResNet18 achieves a 1.60\% increment on SEW ResNet18 and a 2.46\% increment on Spiking ResNet18. The accuracy of our ResNet34 does not exceed SEW ResNet34. However, SEW ResNet34~\cite{2021Deep} transmits information with integers, which is not a typical SNN. For a fair comparison, we also report the result of Spiking ResNet34 in \cite{2021Deep} which is worse than our method. Moreover, our InfLoR-based ResNet34 with 4 timesteps still obviously outperforms STBP-tdBN-based RersNet34 with 6 timesteps.

\begin{table*}[!h]\small
	\centering	
	\caption{Training Spiking Neural Networks on CIFAR10-DVS.}	
	\label{tab:Comparison}	
	\begin{tabular}{lllccc}	
		\toprule
		Dataset & Method & Type & Architecture & Timestep & Accuracy \\	
		\toprule
		\multirow{4}{*}{CIFAR10-DVS}	
		& Rollout~\cite{2020Efficient} & Rollout & DenseNet & 10 & 66.80\%   \\	
		& STBP-tdBN~\cite{2020Going} & SNN training & ResNet19 & 10 & 67.80\%   \\  
		\cline{2-6}
		& \multirow{2}{*}{\textbf{InfLoR}} & \multirow{2}{*}{SNN training} & {ResNet19} 
		& 10 & \textbf{75.50\%}$\pm 0.12$   \\
		&  &  &											                 {ResNet20} & 10 & \textbf{75.10\%}$\pm 0.09$   \\
		\bottomrule				         	
	\end{tabular}	
\end{table*}

\textbf{CIFAR10-DVS.}
For the neuromorphic dataset, CIFAR10-DVS, InfLoR-SNN achieves the best performance with 75.50\% and 75.10\% top-1 accuracy in 10 timesteps with ResNet19 and ResNet18 as backbones, and obtains 7.80\% improvement compared with STBP-tdBN for ResNet19. It's worth noting that, as a more complex model, ResNet19 only performs a little better than ResNet20 on CIFAR10-DVS. It might be that this neuromorphic dataset suffers much more noise than static ones, thus a more complex model is easier to overfit.

\section{Conclusions}

This work aims at addressing the information loss problem caused by the ``Hard Reset" mechanism of neurons and the $0/1$ spike quantification. Then, the SRIF model, which will drive the membrane potential to a dynamic reset potential, and the MPR that can adjust the membrane potential to a new value closer to quantification spikes than itself are proposed. A detailed analysis of why the SRIF and MPR can reduce the information loss is provided. Furthermore, abundant ablation studies of the proposed methods are given. Combining these two methods, our SNNs outperform other state-of-the-art methods.

\clearpage
%
%
\bibliographystyle{splncs04}
\bibliography{egbib}

\begin{thebibliography}{10}
\providecommand{\url}[1]{\texttt{#1}}
\providecommand{\urlprefix}{URL }
\providecommand{\doi}[1]{https://doi.org/#1}

\bibitem{2015TrueNorth}
Akopyan, F., Sawada, J., Cassidy, A., Alvarez-Icaza, R., Arthur, J., Merolla,
  P., Imam, N., Nakamura, Y., Datta, P., Nam, G.J., et~al.: Truenorth: Design
  and tool flow of a 65 mw 1 million neuron programmable neurosynaptic chip.
  IEEE transactions on computer-aided design of integrated circuits and systems
   \textbf{34}(10),  1537--1557 (2015)

\bibitem{2016Simple}
Bewley, A., Ge, Z., Ott, L., Ramos, F., Upcroft, B.: Simple online and realtime
  tracking. In: 2016 IEEE international conference on image processing (ICIP).
  pp. 3464--3468. IEEE (2016)

\bibitem{bu2022optimal}
Bu, T., Fang, W., Ding, J., Dai, P., Yu, Z., Huang, T.: Optimal ann-snn
  conversion for high-accuracy and ultra-low-latency spiking neural networks.
  In: International Conference on Learning Representations (2021)

\bibitem{cubuk2019autoaugment}
Cubuk, E.D., Zoph, B., Mane, D., Vasudevan, V., Le, Q.V.: Autoaugment: Learning
  augmentation policies from data. arXiv preprint arXiv:1805.09501  (2018)

\bibitem{2018Loihi}
Davies, M., Srinivasa, N., Lin, T.H., Chinya, G., Cao, Y., Choday, S.H., Dimou,
  G., Joshi, P., Imam, N., Jain, S., et~al.: Loihi: A neuromorphic manycore
  processor with on-chip learning. Ieee Micro  \textbf{38}(1),  82--99 (2018)

\bibitem{2009ImageNet}
Deng, J., Dong, W., Socher, R., Li, L.J., Li, K., Fei-Fei, L.: Imagenet: A
  large-scale hierarchical image database. In: 2009 IEEE conference on computer
  vision and pattern recognition. pp. 248--255. Ieee (2009)

\bibitem{deng2022temporal}
Deng, S., Li, Y., Zhang, S., Gu, S.: Temporal efficient training of spiking
  neural network via gradient re-weighting. arXiv preprint arXiv:2202.11946
  (2022)

\bibitem{devries2017improved}
DeVries, T., Taylor, G.W.: Improved regularization of convolutional neural
  networks with cutout. arXiv preprint arXiv:1708.04552  (2017)

\bibitem{2015Fastclassifying}
Diehl, P.U., Neil, D., Binas, J., Cook, M., Liu, S.C., Pfeiffer, M.:
  Fast-classifying, high-accuracy spiking deep networks through weight and
  threshold balancing. In: 2015 International joint conference on neural
  networks (IJCNN). pp.~1--8. ieee (2015)

\bibitem{2021Deep}
Fang, W., Yu, Z., Chen, Y., Huang, T., Masquelier, T., Tian, Y.: Deep residual
  learning in spiking neural networks. Advances in Neural Information
  Processing Systems  \textbf{34},  21056--21069 (2021)

\bibitem{2020Incorporating}
Fang, W., Yu, Z., Chen, Y., Masquelier, T., Huang, T., Tian, Y.: Incorporating
  learnable membrane time constant to enhance learning of spiking neural
  networks. In: Proceedings of the IEEE/CVF International Conference on
  Computer Vision. pp. 2661--2671 (2021)

\bibitem{gong2019differentiable}
Gong, R., Liu, X., Jiang, S., Li, T., Hu, P., Lin, J., Yu, F., Yan, J.:
  Differentiable soft quantization: Bridging full-precision and low-bit neural
  networks. In: Proceedings of the IEEE/CVF International Conference on
  Computer Vision. pp. 4852--4861 (2019)

\bibitem{guo2023direct}
Guo, Y., Huang, X., Ma, Z.: Direct learning-based deep spiking neural networks:
  a review. Frontiers in Neuroscience  \textbf{17},  1209795 (2023)

\bibitem{guo2023rmploss}
Guo, Y., Liu, X., Chen, Y., Zhang, L., Peng, W., Zhang, Y., Huang, X., Ma, Z.:
  Rmp-loss: Regularizing membrane potential distribution for spiking neural
  networks. arXiv preprint arXiv:2308.06787  (2023)

\bibitem{guo2023joint}
Guo, Y., Peng, W., Chen, Y., Zhang, L., Liu, X., Huang, X., Ma, Z.: Joint
  a-snn: Joint training of artificial and spiking neural networks via
  self-distillation and weight factorization. Pattern Recognition p. 109639
  (2023)

\bibitem{Guo_2022_CVPR}
Guo, Y., Tong, X., Chen, Y., Zhang, L., Liu, X., Ma, Z., Huang, X.: Recdis-snn:
  Rectifying membrane potential distribution for directly training spiking
  neural networks. In: Proceedings of the IEEE/CVF Conference on Computer
  Vision and Pattern Recognition (CVPR). pp. 326--335 (June 2022)

\bibitem{guo2023membrane}
Guo, Y., Zhang, Y., Chen, Y., Peng, W., Liu, X., Zhang, L., Huang, X., Ma, Z.:
  Membrane potential batch normalization for spiking neural networks. arXiv
  preprint arXiv:2308.08359  (2023)

\bibitem{2020Deep}
Han, B., Roy, K.: Deep spiking neural network: Energy efficiency through time
  based coding. In: European Conference on Computer Vision. pp. 388--404.
  Springer (2020)

\bibitem{2020RMP}
Han, B., Srinivasan, G., Roy, K.: Rmp-snn: Residual membrane potential neuron
  for enabling deeper high-accuracy and low-latency spiking neural network. In:
  Proceedings of the IEEE/CVF conference on computer vision and pattern
  recognition. pp. 13558--13567 (2020)

\bibitem{2016Deep}
He, K., Zhang, X., Ren, S., Sun, J.: Deep residual learning for image
  recognition. In: Proceedings of the IEEE conference on computer vision and
  pattern recognition. pp. 770--778 (2016)

\bibitem{2017Channel}
He, Y., Zhang, X., Sun, J.: Channel pruning for accelerating very deep neural
  networks. In: Proceedings of the IEEE international conference on computer
  vision. pp. 1389--1397 (2017)

\bibitem{2008SpiNNaker}
Khan, M.M., Lester, D.R., Plana, L.A., Rast, A., Jin, X., Painkras, E., Furber,
  S.B.: Spinnaker: mapping neural networks onto a massively-parallel chip
  multiprocessor. In: 2008 IEEE International Joint Conference on Neural
  Networks (IEEE World Congress on Computational Intelligence). pp. 2849--2856.
  Ieee (2008)

\bibitem{Kim2018Deep}
Kim, J., Kim, H., Huh, S., Lee, J., Choi, K.: Deep neural networks with
  weighted spikes. Neurocomputing  \textbf{311},  373--386 (2018)

\bibitem{2019Spiking}
Kim, S., Park, S., Na, B., Yoon, S.: Spiking-yolo: spiking neural network for
  energy-efficient object detection. In: Proceedings of the AAAI conference on
  artificial intelligence. vol.~34, pp. 11270--11277 (2020)

\bibitem{CIFAR-10}
Krizhevsky, A., Nair, V., Hinton, G.: Cifar-10 (canadian institute for advanced
  research). URL http://www. cs. toronto. edu/kriz/cifar. html  \textbf{5}(4),
  ~1 (2010)

\bibitem{2020Efficient}
Kugele, A., Pfeil, T., Pfeiffer, M., Chicca, E.: Efficient processing of
  spatio-temporal data streams with spiking neural networks. Frontiers in
  Neuroscience  \textbf{14}, ~439 (2020)

\bibitem{Ledinauskas2020}
Ledinauskas, E., Ruseckas, J., Juršėnas, A., Burachas, G.: Training deep
  spiking neural networks  (06 2020)

\bibitem{2019Enabling}
Lee, C., Sarwar, S.S., Panda, P., Srinivasan, G., Roy, K.: Enabling spike-based
  backpropagation for training deep neural network architectures. Frontiers in
  neuroscience p.~119 (2020)

\bibitem{2017CIFAR10}
Li, H., Liu, H., Ji, X., Li, G., Shi, L.: Cifar10-dvs: an event-stream dataset
  for object classification. Frontiers in neuroscience  \textbf{11}, ~309
  (2017)

\bibitem{li2021free}
Li, Y., Deng, S., Dong, X., Gong, R., Gu, S.: A free lunch from ann: Towards
  efficient, accurate spiking neural networks calibration. In: International
  Conference on Machine Learning. pp. 6316--6325. PMLR (2021)

\bibitem{li2019additive}
Li, Y., Dong, X., Wang, W.: Additive powers-of-two quantization: An efficient
  non-uniform discretization for neural networks. arXiv preprint
  arXiv:1909.13144  (2019)

\bibitem{li2021brecq}
Li, Y., Gong, R., Tan, X., Yang, Y., Hu, P., Zhang, Q., Yu, F., Wang, W., Gu,
  S.: Brecq: Pushing the limit of post-training quantization by block
  reconstruction. arXiv preprint arXiv:2102.05426  (2021)

\bibitem{li2021differentiable}
Li, Y., Guo, Y., Zhang, S., Deng, S., Hai, Y., Gu, S.: Differentiable spike:
  Rethinking gradient-descent for training spiking neural networks. Advances in
  Neural Information Processing Systems  \textbf{34},  23426--23439 (2021)

\bibitem{2016SGDR}
Loshchilov, I., Hutter, F.: Sgdr: Stochastic gradient descent with warm
  restarts. arXiv preprint arXiv:1608.03983  (2016)

\bibitem{2020Exploring}
Lu, S., Sengupta, A.: Exploring the connection between binary and spiking
  neural networks. Frontiers in Neuroscience  \textbf{14}, ~535 (2020)

\bibitem{2015Darwin}
Ma, D., Shen, J., Gu, Z., Zhang, M., Zhu, X., Xu, X., Xu, Q., Shen, Y., Pan,
  G.: Darwin: A neuromorphic hardware co-processor based on spiking neural
  networks. Journal of Systems Architecture  \textbf{77},  43--51 (2017)

\bibitem{2019Surrogate}
Neftci, E.O., Mostafa, H., Zenke, F.: Surrogate gradient learning in spiking
  neural networks: Bringing the power of gradient-based optimization to spiking
  neural networks. IEEE Signal Processing Magazine  \textbf{36}(6),  51--63
  (2019)

\bibitem{park2019fast}
Park, S., Kim, S., Choe, H., Yoon, S.: Fast and efficient information
  transmission with burst spikes in deep spiking neural networks. In: 2019 56th
  ACM/IEEE Design Automation Conference (DAC). pp.~1--6. IEEE (2019)

\bibitem{2020T2FSNN}
Park, S., Kim, S., Na, B., Yoon, S.: T2fsnn: Deep spiking neural networks with
  time-to-first-spike coding. In: 2020 57th ACM/IEEE Design Automation
  Conference (DAC). pp.~1--6. IEEE (2020)

\bibitem{2019Towards}
Pei, J., Deng, L., Song, S., Zhao, M., Zhang, Y., Wu, S., Wang, G., Zou, Z.,
  Wu, Z., He, W., et~al.: Towards artificial general intelligence with hybrid
  tianjic chip architecture. Nature  \textbf{572}(7767),  106--111 (2019)

\bibitem{2018Model}
Polino, A., Pascanu, R., Alistarh, D.: Model compression via distillation and
  quantization. arXiv preprint arXiv:1802.05668  (2018)

\bibitem{2020DIET}
Rathi, N., Roy, K.: Diet-snn: Direct input encoding with leakage and threshold
  optimization in deep spiking neural networks. arXiv preprint arXiv:2008.03658
   (2020)

\bibitem{2020Enabling}
Rathi, N., Srinivasan, G., Panda, P., Roy, K.: Enabling deep spiking neural
  networks with hybrid conversion and spike timing dependent backpropagation.
  arXiv preprint arXiv:2005.01807  (2020)

\bibitem{2015U}
Ronneberger, O., Fischer, P., Brox, T.: U-net: Convolutional networks for
  biomedical image segmentation. In: International Conference on Medical image
  computing and computer-assisted intervention. pp. 234--241. Springer (2015)

\bibitem{2019Going}
Sengupta, A., Ye, Y., Wang, R., Liu, C., Roy, K.: Going deeper in spiking
  neural networks: Vgg and residual architectures. Frontiers in neuroscience
  \textbf{13}, ~95 (2019)

\bibitem{2018Spatio}
Wu, Y., Deng, L., Li, G., Zhu, J., Shi, L.: Spatio-temporal backpropagation for
  training high-performance spiking neural networks. Frontiers in neuroscience
  \textbf{12}, ~331 (2018)

\bibitem{2018Direct}
Wu, Y., Deng, L., Li, G., Zhu, J., Xie, Y., Shi, L.: Direct training for
  spiking neural networks: Faster, larger, better. In: Proceedings of the AAAI
  Conference on Artificial Intelligence. vol.~33, pp. 1311--1318 (2019)

\bibitem{2020Temporal}
Zhang, W., Li, P.: Temporal spike sequence learning via backpropagation for
  deep spiking neural networks. Advances in Neural Information Processing
  Systems  \textbf{33},  12022--12033 (2020)

\bibitem{2020Going}
Zheng, H., Wu, Y., Deng, L., Hu, Y., Li, G.: Going deeper with directly-trained
  larger spiking neural networks. In: Proceedings of the AAAI Conference on
  Artificial Intelligence. vol.~35, pp. 11062--11070 (2021)

\end{thebibliography}
\end{document}